\def\year{2020}\relax
\documentclass[letterpaper]{article} 
\usepackage{aaai20}  
\usepackage{times}  
\usepackage{helvet} 
\usepackage{courier}  
\usepackage[hyphens]{url}  
\usepackage{graphicx} 
\usepackage{graphicx}  
\usepackage{amsmath}
\title{CorGAN: Correlation-Capturing Convolutional Generative Adversarial Networks for Generating Synthetic Healthcare Records}
\author{\LARGE \textbf{Amirsina Torfi, Edward A.~Fox}\\ \\ 
\Large \textbf{Virginia Polytechnic Institute and State University~(Virginia Tech)}\\ 
Department of Computer Science\\
Blacksburg, Virginia 24061\\
\{atorfi,fox\}@vt.edu 
}

\begin{document}

\maketitle

\begin{abstract}
Deep learning models have demonstrated high-quality performance in areas such as image classification and speech processing.
However, creating a deep learning model using electronic health record (EHR) data, requires addressing particular privacy challenges that are unique to researchers in this domain.
This matter focuses attention on generating realistic synthetic data while ensuring privacy.
In this paper, we propose a novel framework called correlation-capturing Generative Adversarial Network (CorGAN), to generate synthetic healthcare records.
In CorGAN we utilize Convolutional Neural Networks to capture the correlations between adjacent medical features in the data representation space by combining Convolutional Generative Adversarial Networks and Convolutional Autoencoders.
To demonstrate the model fidelity, we show that CorGAN generates synthetic data with performance similar to that of real data in various Machine Learning settings such as classification and prediction.~We also give a privacy assessment and report on statistical analysis regarding realistic characteristics of the synthetic data. The software of this work is open-source and is available at:~\url{https://github.com/astorfi/cor-gan}{https://github.com/astorfi/cor-gan}.
\end{abstract}

\section{Introduction}

Adoption of Electronic Health Records~(EHRs) by the healthcare community, along with the massive quantity of available data, has led to calls for employing promising data-driven methods inspired by Artificial Intelligence (AI).
Data-powered tools alter how clinicians and healthcare bureaus approach and satisfy patients' needs for care.
However, extending EHR adoption to also support data access for research and development purposes, is far from being practical in the healthcare domain, due to privacy restrictions.

De-identification of EHR data is often employed for mitigation of privacy risks.
However, questions and doubts have increased about the safety of prolonged use of de-identification methods regarding their vulnerability to information leakage~\cite{el2011re}. Accordingly, more recent attention has focused on Synthetic Data Generation (SDG) which can satisfy reliably the needs for privacy.

We aim to create realistic synthetic EHR data by Generative Adversarial Networks~(GANs), which have been successfully employed in applications such as image generation~\cite{reed2016generative,brock2018large,karras2018style}, video generation~\cite{vondrick2016generating,tulyakov2018mocogan}, and image translation~\cite{isola2017image,kim2017learning,dong2017unsupervised}.
Contributions of this work include:

\begin{itemize}
    \item We propose an efficient architecture to generate synthetic healthcare records using Convolutional GANs and \textit{Convolutional Autoencoders}~(CAs) which we call \textit{``CorGAN''}. We demonstrate that CorGAN can effectively generate both discrete and continuous synthetic records.
    \item We demonstrate the effectiveness of utilizing Convolutional Neural Networks~(CNNs) as opposed to Multilayer Perceptrons to capture inter-correlation between features.
    \item We show that CorGAN can generate realistic synthetic data that performs similarly to real data on classification tasks, according to  our analysis and assessments.
    \item We report on a privacy assessment of the model and demonstrate that CorGAN provides an acceptable level of privacy, by varying the amount of synthetically generated data and amount of data known to an adversary.
\end{itemize}

\section{Related Works}\label{sec:RelatedWorks}

Some distinguished efforts were conducted in a variety of domains about synthetic data generation~\cite{walonoski2017synthea,buczak2010data,mclachlan2016using,park2013perturbed}.
But some of these works are overly disease-specific, unrealistic, or have failed to provide any substantial measurements regarding privacy.

Highly relevant is ``medGAN''~\cite{choi2017generating}, using GANs for synthetic discrete EHR data.
But in contrast to medGAN, we consider the temporal nature of the data and local correlation between features.
Instead of regular multi-layer perceptrons, we base our architecture on CNNs and provide empirical results to demonstrate the superior performance in capturing inter-correlations between data features.

\section{Method}\label{sec:Method}

\subsection{Discrete EHR Data Description}\label{sec:Methodsub:Data Description}

Many discrete variables (e.g., diagnosis, procedure code) are available in the dataset.
Let's assume there are $\boldsymbol{|D|}$ discrete variables and the vector $\boldsymbol{V_C \in {N_0 }^{|D|}}$~(where $N_0$ indicates natural numbers including zero) is in a vector space.
The $j^{th}$ dimension designates the number of incidents of the $j^{th}$ variable in a subject's medical records.
We can represent a patient's visit~(encounter event) by a binary vector $\boldsymbol{V_B \in {\{0,1\} }^{|D|}}$, where the $j^{th}$ dimension shows whether the $j^{th}$ variable occurred in the patient record.
We represent the input space as a matrix in which columns indicate discrete variables in the EHR record. Such representation extracts multiple patients' records representing different points in time.

\subsection{Generative Adversarial Networks}\label{sec:Methodsub:Preliminary}

A \textit{Generative Adversarial Network~(GAN)}, introduced in \cite{goodfellow2014generative}, is a combination of two neural networks, a discriminator and a generator.
The whole network is trained in an iterative process.
First the generator network produces a fake sample.
Then the discriminator network tries to determine whether this sample (ex.: an input image) is real or fake, i.e., whether it came from the real training data.
The goal of the generator is to fool the discriminator so it believes the artificial (i.e., generated) samples synthesized by the generator are real.

The generator goal is to learn the distribution $p_g$ over data $\textbf{x}$.~In that regard,~$p_z(\textbf{z})$ represents the input noise variables distribution which generates random data shown by $G(\textbf{z}; \theta_g)$.
The function $G$ is differentiable with parameters $\theta_g$.
The discriminator, $D(\textbf{x}; \theta_d)$, decides if its input data is real or fake.
$D$ is trained to distinguish the training samples from $G$ by minimizing $ log(1 - D(G(z))$).
$D$ and $G$ perform the following min-max game with value function V(G,~D):

\begin{equation}\label{eq:GAN}
\begin{split}
\underset{G}{Min}\underset{D}{Max}V(G,D) = E_{x\sim p_{data}(x)}[logD(x)] + \\ E_{z\sim p_{z}(z)}[1-logD(G(z))]
\end{split}
\end{equation}

\subsection{Proposed Architecture}\label{sec:Methodsub:ProposedArchitecture}

\begin{figure}
\centering
    \includegraphics[width=.95\columnwidth]{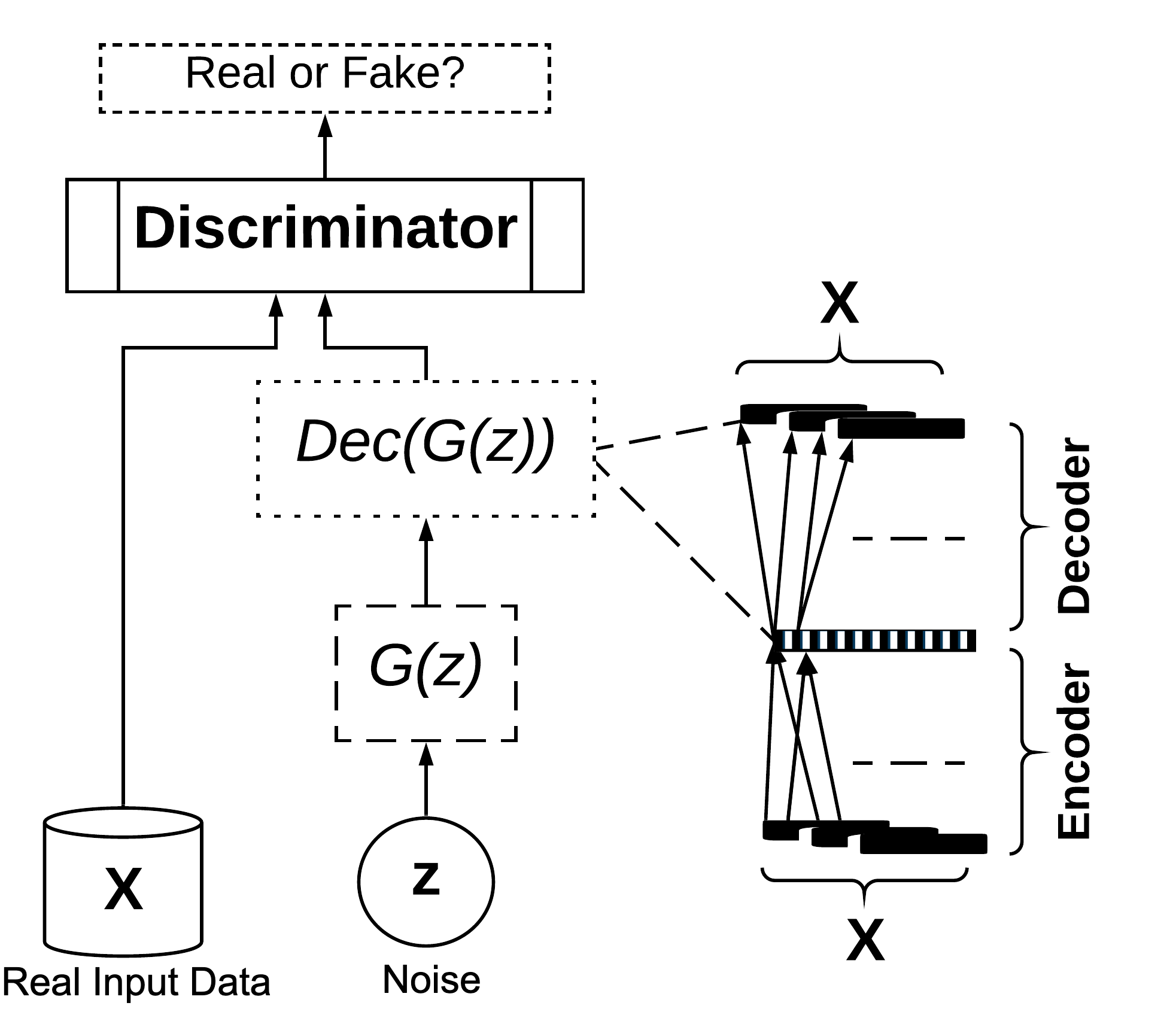}
    \caption[The proposed architecture for synthetic data generation]{The architecture for generating synthetic data from real samples. The right side of the figures, shows the pretrained convolutional autoencoder which its decoder part is being used to transform the generated continuous samples to their discrete equivalents.} \label{fig:proposedarch}
\end{figure}

We use the architecture in Fig.~\ref{fig:proposedarch}.
The discrete input $\boldsymbol{X}$ represents the source EHR data;~$\boldsymbol{z}$ is the random distribution for the generator $\boldsymbol{G}$; $\boldsymbol{G}$ is the employed neural network architecture; $\boldsymbol{Dec(G(z))}$ refers to the decoding function which is used to transform the generator $\boldsymbol{G}$ continuous output to their equivalent discrete values.~The discriminator $\boldsymbol{D}$ attempts to distinguish real input $\boldsymbol{X}$ from the discrete synthetic output $\boldsymbol{Dec(G(z))}$.~For the generator and the discriminator,~a 1-Dimensional Convolutional GAN architecture is utilized.

Consider the decoding function $\boldsymbol{Dec(.)}$.
GANs are known for generating continuous values and encountering trouble when dealing with discrete variables.
Recently, researchers proposed solutions to the problem of generating discrete variables~\cite{hjelm2017boundary,wang2017irgan,kim2017adversarially,yu2017seqgan}.
Some approaches use the indirect method such that they create a separate model to transform continuous to discrete data~\cite{choi2017generating}.
Regarding EHR data generation, we are dealing with discrete data.
Hence,~our generative model should create discrete data directly,~or there should be a function to transform the continuous data samples into discrete equivalents.
We chose the second approach, and employed autoencoders.

Considering Fig.~\ref{fig:proposedarch}, the autoencoder digests (right part of the figure) discrete values and reconstructs the same discrete values as well.
The autoencoder structure consists of two main elements: encoder and decoder.
While encoding, the autoencoder transforms the discrete space into a corresponding (we call it equivalent as well) continuous space (the output of the hidden layer) and the decoder reverses the process.
The \textit{Binary Cross-Entropy~(BCE)} loss function is used for training the autoencoder:

\begin{align}\label{eq:binarycrossentropy}
& BCE_{loss}= -\frac{1}{N}\sum_{i=1}^{N}x_ilog(y_i)+(1-x_i)log(1-y_i) \\ 
 & y_i =\boldsymbol{Dec}(\boldsymbol{Enc}(x_i))
\end{align}

We used denoising autoencoders~\cite{vincent2010stacked} to create a more robust pretrained model as we do not expect our model to always generate perfect discrete samples.
After training the autoencoder, we need to use its decoder to convert continuous values to their associated discrete values.
The cost function to train our proposed architecture is similar to Eq.~\ref{eq:GAN} with the exception of operating the decoder on top of the generator.

\begin{equation}\label{eq:GANproposed}
\begin{split}
\underset{G}{Min}\underset{D}{Max}V(G,D) = E_{x\sim p_{data}(x)}[logD(x)] + \\ E_{z\sim p_{z}(z)}[1-logD(\boldsymbol{Dec}(G(z)))]
\end{split}
\end{equation}

As we are dealing with 1D data,
\textit{we chose the 1-Dimensional
Convolutional Autoencoders (1D-CAEs) as a particular form of the regular CAEs}.
This approach enables us to capture the neighboring feature correlations.~We call our proposed architecture \textit{CorGAN}.~It is worth noting that for our experiments with discrete variables, we round the values of $Dec(G(z))$ to their nearest integers (the outcome is zero or one) to guarantee that we train and evaluate the discriminator on discrete values.

\subsection{Training Augmentation}\label{sec:Methodsub:TrainingAugmentation}

One of the primary crash forms for GAN is for the generator to collapse to a set of parameters and always generate the same sample.
This phenomenon is called \textit{Mode Collapse}. Some approaches have been proposed to handle the mode collapse issue such as \textit{minibatch discrimination}~\cite{salimans2016improved} and \textit{unrolled GANs}~\cite{metz2016unrolled}.~We utilized minibatch discrimination due to its better stability. We also utilized batch normalization~\cite{ioffe2015batch} to improve the generator's learning abilities. Furthermore, we used LeakyRelu activation unit \cite{maas2013rectifier} as it consistently demonstrated equal or better results over other common activation functions~\cite{xu2015empirical}.




\subsection{Privacy}\label{sec:Methodsub:Privacy}


We utilize the \textit{Membership Inference (MI)} attack as an approach to measure the privacy.~\textit{Membership Inference~(MI)}, proposed in~\cite{shokri2017membership}, refers to determining whether a given record generated by a known machine learning model was used as part of the training data.~If the adversary has complete access to the records of a particular patient and can recognize their employment in the model training, that is an indication of information leakage, as it can jeopardize the whole dataset privacy or at least the particular patient's private information.~Here, we will assume the adversary \textit{has the synthetically generated data as well as a portion of the compromised real data}.



\section{Experiments}\label{sec:Experiments}

We evaluated \textit{CorGAN} with two datasets.
First, we explain the datasets and baseline models.
Then, we provide the results regarding the evaluation of the synthetic data in terms of the realistic characteristics. 
Finally, we report on a privacy assessment of the model.

\subsection{Datasets}\label{sec:Experimentssub:Dataset}

We used two publicly available datasets in this study.
The first is the \textit{MIMIC-III} dataset~\cite{johnson2016mimic} consisting of the medical records of almost 46K patients.
From MIMIC-III, we extracted ICD-9 codes only.
We represent a patient record as a fixed-size vector with 1071 entries for each patient record.
This dataset is used for experiments with binary discrete variables.

We conducted our experiments regarding continuous variables with the \textit{UCI Epileptic Seizure Recognition} dataset~\cite{andrzejak2001indications}.
This dataset characterizes brain activities.
The core task is classification, regarding if a sample indicates a seizure activity.
The number of features and samples are 179 and 11500, respectively.
Almost 20\% of the samples are categorized as seizure activity.
So, we are dealing with an unbalanced dataset in a binary classification setting.
The first 178 features are the values of the Electroencephalogram~(EEG) recordings at different time points, and the last feature is the class.
There are five values for the class label ($y={1,...,5}$).
Except for $y=1$, the rest of the classes indicate subjects who do not have an epileptic seizure.
Dataset statistics are given in Table~\ref{tab:uci_stats}.


\begin{table}[t]
\caption{Statistics of the UCI Epileptic Seizure Recognition dataset.}
\centering
\resizebox{.95\columnwidth}{!}{
\begin{tabular}{|l|c|c|c}
\hline
\textbf{Dataset} & \textbf{UCI}\\
\hline
\# of patients & 500\\ 
Each patient's data points & 4097\\
Each patient's duration of recording & 23.5 seconds\\
\# data points chunks per patient & 23\\
\# of data points per chunk & 178\\
Duration per chunk & 1 second\\
Data type & Continuous EEG\\
\hline
\end{tabular}}
\label{tab:uci_stats}
\end{table}

\subsection{Models}\label{sec:Experimentssub:Models}

To show the effectiveness of our proposed architecture, we compare our results with different baseline methods as below:

\begin{itemize}
\item \textbf{Stacked Deep Boltzmann Machines (DBMs):} We trained a stacked Deep Boltzmann Machine~(DBM) \cite{hinton2009replicated}. After which, we used Gibbs sampling to generate synthetic binary samples. All hidden layers have 256 dimensions. We employed greedy contrastive divergence to create the model.~We ran Gibbs sampling for 500 iterations per sample.

\item \textbf{Variational Autoencoder (VAE):} We used VAEs~\cite{kingma2013auto} as one of our baseline models.
For both the encoder and the decoder, we used 1D convolutional neural networks, each having two hidden layers.
All hidden layers have the size of 128.
We trained VAE with Adam optimizer for 500 epochs and for the batch size of 500.

\item \textbf{medGan:} The medGan \cite{choi2017generating} architecture consists of the following elements; 
\textbf{(1)} regular multilayer perceptrons for autoencoder, discriminator, and generator.~\textbf{(2)} shortcut connections to improve the power of generator.~\textbf{(3)} minibatch-averaging~\cite{choi2017generating} to cope with the mode collapse.

\end{itemize}

\section{Evaluation}\label{sec:Evaluation}

In this section, we report our evaluation results regarding the quality of the synthetic data and the privacy risks. Here, we divide the dataset into a training $\mathcal{S}_{tr} \in \{0,1\}^{R \times |\mathcal{M}|}$ and a test set $\mathcal{S}_{te} \in \{0,1\}^{T \times |\mathcal{M}|}$, where $|\mathcal{M}|$ is the feature size and is consistent for all sets. We use $\mathcal{S}_{tr}$ to train the models, then generate synthetic samples $\mathcal{S}_{syn} \in \{0,1\}^{S \times |\mathcal{M}|}$ using the trained model.~Noted that we usually use the same number of samples for $\mathcal{S}_{syn}$ and $\mathcal{S}_{tr}$.

\subsection{Evaluation of the Synthetic Data Quality}\label{sec:Experimentssub:realistic}

We use the following two metrics to evaluate our synthetically generated data.

\begin{itemize}
    \item  \textbf{Dimension-wise probability:} As a basic sanity check to see if our proposed models learned the distribution of the real data (for each dimension), we report the dimension-wise probability. This measurement refers to the Bernoulli success probability of each dimension~(each dimension is a unique ICD-9 code).
    
    \item  \textbf{Dimension-wise prediction:} This approach measures how robust the model catches the inter-dimensional connections of the real data samples. Assume $\mathcal{S}_{tr}$ is used to generate $\mathcal{S}_{syn}$.~Then, one random fixed dimension~($k$) from each $\mathcal{S}_{syn}$ and $\mathcal{S}_{tr}$ are selected as $\mathcal{S}_{syn,k} \in \{0,1\}^{N \times 1}$ and $\mathcal{S}_{tr,k} \in \{0,1\}^{N \times 1}$. We call it the testing dimension. The rest of the dimensions~($\mathcal{S}_{syn, \backslash k} \in \{0,1\}^{N \times 1}$ and $\mathcal{S}_{tr, \backslash k} \in \{0,1\}^{N \times 1}$) are used to train a classifier, which aims to predict the value of the testing dimension of the test set $\mathcal{S}_{te,k} \in \{0,1\}^{N \times 1}$.
    
    \item  \textbf{Binary Classification:} We use this metric for our experiments with continuous data.~To empirically verify the quality of the synthetic data, we consider two different settings.~\textbf{(A)} Train and test the predictive models on the real data.~\textbf{(B)} train the predictive model on synthetic data and test it on the real data.~If the model evaluated in setting (B), represents competitive results with the same model performed in setting (A), we can conclude the synthetic data has good predictive modeling similar to the real data.
\end{itemize}{}

\begin{figure*}
\centering
    \includegraphics[width=.95\textwidth]{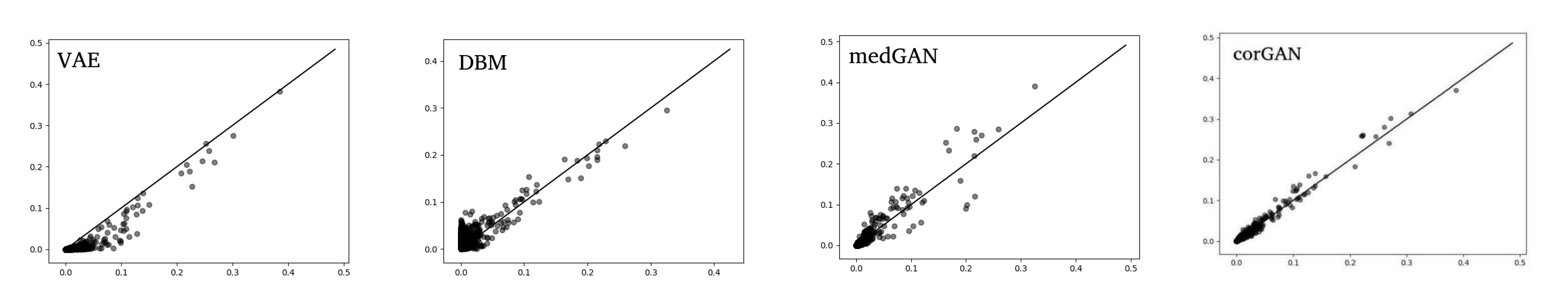}
    \caption[]{he scatter plots of dimension-wise probability. Each point depicts one of 1071 unique diagnosis codes. The x-axis and y-axis represent the Bernoulli success probability for real and synthetic datasets, respectively. The diagonal line shows the ideal case.} \label{fig:dwpmimic}
\end{figure*}

For Dimension-wise probability and Dimension-wise prediction experiments, we used \textit{MIMIC-III} dataset and for Binary Classification experiments we used \textit{UCI Epileptic Seizure Recognition} dataset.~The results regarding the investigation of dimension-wise probability are depicted in Fig.~\ref{fig:dwpmimic}. As can be seen, the \textit{CorGAN} is superior compared to other methods. An interesting observation is that the VAE is never generating any synthetic data for which the probability of occurrence of a diagnosis code is higher than its counterpart in the real data.

For dimension-wise prediction~(Table~\ref{table:dwpredmimic}), we use the following classifiers the predictive model types:~Logistic Regression,~Random Forests~\cite{breiman2001random},~Linear SVM~\cite{cortes1995support},~and Decision Tree~\cite{quinlan1986induction}.~For our experiments, we conduct E=100 number of runs.~In each run, we pick a random testing dimension from test set ($\mathcal{S}_{syn,k} \in \{0,1\}^{N \times 1}$) and will train each predictive model on $\mathcal{S}_{syn, \backslash k}$ and $\mathcal{S}_{tr, \backslash k}$. This results in having two models as $Model^{type}_{syn}$ and $Model^{type}_{real}$. The superscript refers to the kind of model that was used to be trained on both real and synthetic sets.~We then report the performance over all predictive models and for all experiments using the F1-score variation. F1-score variation means the difference between the F-1 score obtained from training the model on synthetic and real datasets. 

For the MIMIC dataset experiments, although the temporal information is ignored, the medical health diagnosis are sorted in terms of similarity in the feature vector of the MIMIC data. Therefore, 1D CNNs capture the correlation between features rather than the temporal information. 


\begin{table}[t]
\caption{Comparison of different baseline architectures. The reported metric demonstrate the mean and standard deviation of the F-1 score differences. A better model has a closer score to zero.}\smallskip
\centering
\resizebox{.95\columnwidth}{!}{
\begin{tabular}{|c|c|}
\hline
Generative Model & F1-Score\\
\hline
DBM~\cite{hinton2009replicated} & 0.12 $\pm$ 0.052\\
VAE~\cite{kingma2013auto} & 0.069 $\pm$ 0.043\\
medGAN~\cite{choi2017generating} & 0.043 $\pm$ 0.049\\
\textbf{CorGAN}~[ours] & \textbf{0.021} $\pm$ \textbf{0.045}\\
\hline
\end{tabular}}
\label{table:dwpredmimic}
\end{table}

For binary classification experiments,~we used the same predictive models as for the dimension-wise predictions. We reported the averaged AUROC~(averaged area under the ROC for all models) and AUPRC~(averaged area under the PR curve for all models) for the models' evaluations.~The difference with our experiments here is that we are not dealing with binary variables.~Hence, for the medGAN and CorGAN methods we eliminate the autoencoder as depicted in Fig.~\ref{fig:proposedarch}.~As can be observed in Table.~\ref{table:bcuci}, our proposed method outperforms the other methods. As the \textit{UCI Epileptic Seizure Recognition} dataset features contain termporal information, our method is able to capture temporal data information more effectively due to the usage of 1D CNNs.

\begin{table}[t]
\caption{Comparison of different generative models for binary classification. The averaged AUROC and AUROC for utilizing the predictive models on the real data are \textbf{0.95} and \textbf{0.46}.}
\centering
\resizebox{.95\columnwidth}{!}{
\begin{tabular}{|c|c|c|}
\hline
Generative Model & AUROC & AUPRC\\
\hline
DBM & $0.81 \pm 0.017$ & $0.27 \pm 0.013$\\
VAE & $0.84 \pm 0.021$ & $0.31 \pm 0.022$\\
medGAN & $0.89 \pm 0.023$ & $0.35  \pm 0.014$\\
\textbf{CorGAN}~[ours] & \textbf{0.92 $\pm$ 0.012} & \textbf{0.41 $\pm$ 0.015}\\
\hline
\end{tabular}}
\label{table:bcuci}
\end{table}

\subsection{Privacy Assessment}\label{sec:Experimentssub:privacy}

In this section, the experiments are conducted on the MIMIC-III dataset regarding the membership inference attack. For privacy assessment, we randomly take $\mathcal{P}$ samples from each $\mathcal{S}_{tr}$ and $\mathcal{S}_{te}$ and call them $\mathcal{S}^{\mathcal{P}}_{tr}$ and $\mathcal{S}^{\mathcal{P}}_{tr}$. We assume the attacker has the complete knowledge of both $\mathcal{S}^{\mathcal{P}}_{tr}$ and $\mathcal{S}^{\mathcal{P}}_{te}$. Clearly, $\mathcal{S}^{\mathcal{P}}_{tr}$ was used to train the generating model, but $\mathcal{S}^{\mathcal{P}}_{te}$ wasn't. So we have $\mathcal{R} = 2 \times \mathcal{P}$ records. Then,~we compared each of these records with the synthetically generated data samples $\mathcal{S}_{syn} \in \{0,1\}^{S \times |\mathcal{M}|}$.

We compared each of the samples in the set of $\mathcal{S}^{\mathcal{P}}_{te}+\mathcal{S}^{\mathcal{P}}_{tr}$ with each samples in the set of $\mathcal{S}_{syn}$ and we calculate cosine similarity score. Cosine similarity is used since it provides a more meaningful correlation metric as opposed to distance metrics~\cite{mateo2004outlier} used in previous research efforts~\cite{choi2017generating}.~If the score is higher than a threshold,~then it flags the match,~otherwise, we call it a mismatch. For threshold, we randomly select 100 threshold values from a Gaussian distribution with a mean of 0.5 and a standard deviation of 0.01~(ignoring possible negative values), and we report the results which demonstrate the best adversary attack.

\begin{figure}
\centering
    \includegraphics[width=.95\columnwidth]{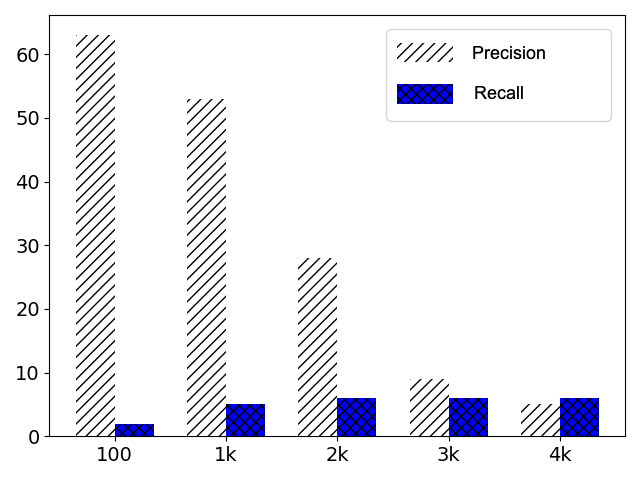}
    \caption[]{The recall/precision as a function of the number of generated
synthetic records.~The number of records known to the adversary is considered fixed and is equal to 100.} \label{fig:prsyntheticnumber}
\end{figure}

For evaluation,~we use \textit{precision} and \textit{recall} metrics.~We conduct two sets of experiments here: \textbf{(1)} investigating the effect of the number of records known by the attacker~(Table.~\ref{table:prnumrecords}) and \textbf{(2)} examining the effect of synthetic data volume on the privacy risk~(Fig.~\ref{fig:prsyntheticnumber}).



As can be seen in Table.~\ref{table:prnumrecords},~by increasing the number of the real patient records known to the adversary,~the attack will be even less accurate. It also demonstrates the fact that higher precision is possible at lower recall rates when the number of known records is not high.~However,~as is evident,~a higher amount of revealed data increases the privacy risk significantly. 

Regarding the effect of number of generated synthetic data on the privacy risk, as can be observed in Fig.~\ref{fig:prsyntheticnumber},~the increasing number of synthetic records does not have a significant effect on the recall,~but it causes a dramatic decrease in precision.~Henceforth,~by having a fixed amount of known records,~a higher number of synthetic patient's records can be very misleading for the adversary.~This empirical observation indicates that the increasing the number of synthetic records, with a fixed number or revealed patient's records to the attacker, does not necessarily raise privacy risk.

\begin{table}[t]
\caption{The precision and recall demonstrated as a function of the number of patients whose data is revealed to the attacker. $\mathcal{U}$ = \# of Known Records to the attacker.}\smallskip
\centering
\resizebox{.95\columnwidth}{!}{
\begin{tabular}{|c|c|c|c|c|c|c|}
\hline
$\mathcal{U}$ & 100 & 1k & 2k & 3k & 4k & 5k\\
\hline
Precision~ & 0.60 & 0.51 & 0.41 & 0.40 & 0.40 & 0.39\\
Recall & 0.05 & 0.10 & 0.19 & 0.28 & 0.27 & 0.28\\
\hline
\end{tabular}}
\label{table:prnumrecords}
\end{table}

\section{Conclusion}\label{sec:Conclusion}

In this work, we proposed CorGAN, which utilizes the convolutional generative adversarial networks to learn the distribution of real patient records. Through precise evaluation using real and synthetic datasets, CorGAN demonstrated decent results for both discrete and continuous records.  We empirically proved the superiority of CNNs over MLPs to capture the correlated features. We believe our method can be effectively extended and employed to longitudinal records as well for which the goal is to capture the temporal characteristics of the data.

\bibliography{main}
\bibliographystyle{aaai}

\end{document}